\documentclass[fleqn,10pt]{wlscirep}
\usepackage[utf8]{inputenc}
\usepackage[T1]{fontenc}
\usepackage{algorithm}
\usepackage{algorithmic}
\usepackage{wrapfig}
\usepackage{amsmath}
\usepackage{graphicx}
\usepackage{subcaption}
\usepackage[active]{srcltx}
\usepackage{color}
\usepackage{lineno}
\usepackage{dirtytalk}
\usepackage{amsthm}
\usepackage{multirow}
\usepackage{comment}
\usepackage{booktabs}
\usepackage{lineno}
\usepackage{adjustbox}

    \newcommand{\removeme}[1]{}

\title{Global Lightning-Ignited Wildfires Prediction and Climate Change Projections based on Explainable Machine Learning Models}

\author[1,*]{Assaf Shmuel}
\author[2,3]{Teddy Lazebnik}
\author[1]{Oren Glickman}
\author[4]{Eyal Heifetz}
\author[4]{Colin Price}
\affil[1]{Department of Computer Science, Bar Ilan University, Ramat Gan, Israel}
\affil[2]{Department of Mathematics, Ariel University, Ariel, Israel}
\affil[3]{Department of Cancer Biology, Cancer Institute, University College London, London, UK}
\affil[4]{Porter School of the Environment and Earth Sciences, Tel Aviv University, Tel Aviv, Israel}

\affil[*]{Corresponding author: assafshmuel91@gmail.com}

\begin{abstract}
Wildfires pose a significant natural disaster risk to populations and contribute to accelerated climate change. 
As wildfires are also affected by climate change, extreme wildfires are becoming increasingly frequent. 
Although they occur less frequently globally than those sparked by human activities, lightning-ignited wildfires play a substantial role in carbon emissions and account for the majority of burned areas in certain regions. 
While existing computational models, especially those based on machine learning, aim to predict lightning-ignited wildfires, they are typically tailored to specific regions with unique characteristics, limiting their global applicability.
In this study, we present machine learning models designed to characterize and predict lightning-ignited wildfires on a global scale. Our approach involves classifying lightning-ignited versus anthropogenic wildfires, and estimating with high accuracy the probability of lightning to ignite a fire based on a wide spectrum of factors such as meteorological conditions and vegetation. Utilizing these models, we analyze seasonal and spatial trends in lightning-ignited wildfires shedding light on the impact of climate change on this phenomenon. We analyze the influence of various features on the models using eXplainable Artificial Intelligence (XAI) frameworks. Our findings highlight significant global differences between anthropogenic and lightning-ignited wildfires. Moreover, we demonstrate that, even over a short time span of less than a decade, climate changes have steadily increased the global risk of lightning-ignited wildfires. This distinction underscores the imperative need for dedicated predictive models and fire weather indices tailored specifically to each type of wildfire.
\end{abstract}
\begin{document}

\flushbottom
\maketitle

\thispagestyle{empty}

\section*{Introduction}

Wildfires are among the most hazardous natural disasters on Earth. Although studies have found a decrease in global burned areas due to anthropogenic activity \cite{Andela}, the frequency of extratropical lightning-ignited wildfires appears to be on the rise as the Earth's climate changes \cite{Janssen}. For example, the frequency of extreme lightning-ignited wildfires has drastically increased in California in the last decades \cite{california}. Lightning is the main cause for wildfire activity in high latitudes both in terms of wildfire occurrence and in terms of burned areas \cite{Canada}. This trend is of considerable concern, as extratropical forests are responsible for a substantial portion of carbon emissions. Lightning-ignited wildfires differ from anthropogenic wildfires in several aspects; for example, lightning-ignited wildfires tend to ignite in remote locations in which firefighters have difficulty extinguishing them before they evolve into extreme dimensions. These fires also tend to ignite in clusters \cite{Characterizing_lightning_wildfires} while forest managers are less likely to fight these fires if they are not impacting settlements and infrastructures. While recent studies have developed machine learning (ML) models to predict wildfires on a global scale \cite{Shmuel}, the different nature of lightning-ignited wildfires requires dedicated models to predict and analyze them separately.

Wildfires can be ignited by cloud-to-ground (CG) lightning strikes. As lighting ignitions occur in thunderstorms, they are often accompanied by precipitation \cite{Kalashnikov}. However, lightning strokes that occur with little precipitation are defined as "dry lightning" and are more likely to cause an ignition. While a common definition for "dry lightning" is less than 2.5 mm of daily precipitation, some studies suggest that the risk of ignition is more complex and depends on additional factors \cite{Kalashnikov}. An additional unique characteristic of lightning ignitions is the phenomenon of holdover wildfires, in which the ignition causes smoldering which can last for several days before a wildfire is detected. Although there exist some reports of extreme cases in which the smoldering phase lasted for several weeks, in the vast majority of cases they are limited to several days or a week \cite{Schultz}.

Recently, global lightning-ignited wildfire analysis and models obtained much attention. For example, a recent study used the EMAC model, a numerical model that aims to describe tropospheric and middle atmosphere processes and their interaction with oceans, land, and influences coming from anthropogenic emissions\cite{compare_2}. The authors used global data from 2009 to 2011 with \(2.8 \times 2.8\) degree and 12 minutes spatial and temporal resolution, respectively. Based on this model, they predict a 41\% increase in global lightning flash rate. They also apply a simple model of wildfire ignition risk (lightning frequency divided by precipitation) and predict an increase in most regions around the world.

In a similar manner, researchers used the Max Planck Institute Earth System Model to explore how changes in lightning induced by climate change alter wildfire activity \cite{compare_1}. To be exact, the authors used the popular ECHAM6 simulator, which allowed them to explore different realistic atmospheric conditions, together with the JSBACH land surface vegetation model, which allows them to represent a spatio-temporal vegetation distribution. Using this setup and empirical values from the literature to obtain geophysical relevant values for the models' parameters, the authors show a non-linear correlation between burned areas and cloud-to-ground lightning activity. Importantly, this approach, where a geophysical simulator is used does not allow the applied generalization of the phenomenon for future events as it allows us only to study the dynamics of the phenomenon and can not take into consideration hidden factors available in real-world data which are not taken into account during the simulators' development.

To address this limitation, several studies adopted the ML model approach. Most studies in this field develop ML models on a regional scale. For example, Vecín-Arias et al. (2016) develop a Random Forest model for the Iberian Peninsula \cite{Vecín-Arias}; Malik et al. (2021) develop an Adaboost model for California \cite{Malik}; Sayad et al. (2019) develop Artificial Neural Network and Support Vector Machine models to predict wildfire risk in Canada \cite{Sayad}. Several studies have also leveraged eXplainable Artificial Intelligence (XAI) frameworks to obtain a better understanding of the models' predictions. XAI is an important field of research which aims to develop models which can be better understood by humans \cite{xai1}. XAI approaches vary between models which are inherently explainable, and methods whose goal is to provide explanations of black-box models \cite{xai2}. The application of the latter in wildfire research is becoming increasingly frequent, with the use of SHapley Additive exPlanations (SHAP) as the most common approach \cite{fireXAI1,fireXAI2,fireXAI3,fireXAI4,fireXAI5,fireXAI6}. Other studies also leverage the intrinsic feature importance ordering by tree ensemble models such as XGBoost \cite{feature_importance_trees,FIT1,FIT2}. While regional models may better capture regional characteristics, their performance when applied to out of distribution regions is mostly low. In addition, effective training of ML models requires a very large number of observations, which is not available on a regional scale.

Few studies have tackled this challenge on a global scale. For instance, Janssen et al. (2023) integrated fire-caused reference, lightning, burned-area, low-impact land, intact-forest, fire-related forest loss, and carbon combustion data from multiple sources to take into consideration the factors known to play a role in lightning-ignited wildfires \cite{Janssen}. Based on this data, the authors trained two XGBoost models \cite{xgboost}. Their models predict the most likely ignition source at a 0.5-degree and monthly resolution. This novel study sheds light on the global distribution of anthropogenic versus lightning-caused wildfires, but does have some limitations. Namely, it is performed in a monthly resolution, which does not allow to verify that the lightnings occured before the ignition; the data used for training are limited to 7 regions; finally, the study classifies wildfire types in hindsight and does not allow prediction of future ignition risk which is required for most real-world applications. Comparably, Coughlan et al. (2021) combined lightning presence, lightning–ignited wildfires, and environmental predictors such as soil water and temperature data \cite{compare_4}. The authors used this data to test Decision Tree, AdaBoost, and Random Forest models \cite{rf,dt,Ada}, together with the recursive feature elimination selection method, for binary classification of lighting-integrated wildfires. While their approach is novel, the trained models reached an accuracy score of 71\% which is not always sufficient in real-world applications.

Following these attempts, there is a gap in the literature for a model that is able to globally predict lightning-ignited wildfires with both a high level of accuracy and spatiotemporal stability. In this study, we build on global datasets of lightning \cite{thunderhours} and fire activity \cite{arteswildfires} to develop an ML model that predicts the conditions in which lightning is expected to ignite a wildfire. We test the developed model both by using a random train-test split of the years, and by testing the year 2021 as a holdout year to demonstrate the generalization capabilities of the model. The best-performing model demonstrates high predictive performance with accuracy exceeding 90\%. We analyze the different characteristics of anthropogenic and lightning-caused wildfires, and demonstrate that models developed for the former perform poorly for the latter and vice versa, emphasizing the need for dedicated models and fire weather indices for lightning-caused wildfires. Finally, we use the developed models to estimate the mean annual increase in lightning ignition risk throughout the data's timespan, and also evaluate the predictions of these models in future climate conditions to estimate the expected trends of lightning-ignited wildfire activity. We find that a concerning increase in lightning ignition risk is already occurring and is likely to continue in the future.

The rest of the paper is organized as follows. First, we present the results of the study, starting with the dynamics analysis of the lightning-ignited wildfires, followed by the proposed model's performance, the anthropogenic versus lightning-ignited wildfires, and the effect of climate change on these dynamics. Next, we discuss the geophysical interpretation of the obtained results and how policymakers can use the proposed model to design better policies. Finally, we formally present the methods and materials used as part of this study, including the datasets, the proposed model's development, and the experimental design.

\section*{Results}
In this section, we outline the obtained results, divided into four main outcomes: an analysis of wildfire occurrence dynamics, the performance of the obtained ML model, a comparison between models of anthropogenic and lightning-ignited wildfires, and climate change trajectory prediction based on the obtained model. Table \ref{tab:data} summarizes the variables used in the study and their sources, following a data pre-processing and merging, as described in detail in the \say{Methods and Materials} section.

\begin{table}[]
\centering
\caption{Summary of the variables used in this study (and by the proposed model) together with their sources.}
\label{tab:data}
\resizebox{\columnwidth}{!}{%
\begin{tabular}{ccccc}
\hline
Group & Variable & Acronym & Used in Model & Source \\ \hline
\multirow{4}{*}{wildfire data} & lightning resulted in ignition (target feature) & burned & v & \multirow{4}{*}{\cite{arteswildfires}} \\ 
 & total burned in fire & total\_burned & x &  \\ 
 & area burned in first day of fire & first\_day\_burned & x &  \\ 
 & fire duration & duration & x &  \\ \hline
thunder data & thunderhours & thunderhours & x & \cite{thunderhours} \\ \hline
\multirow{13}{*}{\begin{tabular}[c]{@{}c@{}}vegetation,\\ meteorological,\\ and anthropogenic\\ factors\end{tabular}} & relative humidity & RH & v & \multirow{11}{*}{\cite{era5}} \\ 
 & temperature & t & v &  \\ 
 & precipitation & prec & v &  \\ 
 & days since last precipitation & days\_since\_prec & v &  \\ 
 & monthly precipitation & monthly\_prec & v &  \\ 
 & wind velocity & v\_total & v &  \\ 
 & wind direction & v\_direction & v &  \\ 
 & low vegetation cover & low\_veg & v &  \\ 
 & high vegetation cover & high\_veg & v &  \\  
 & Normalized Difference Vegetation Index & NDVI & v & \cite{NDVI} \\ 
 & soil moisture & sm & v & \cite{era5} \\ 
 &  vegetation water content (skin reservoir) & water & v & \cite{era5} \\ 
 & population density & pop & v & \cite{populationdata} \\ \hline
historical fire data & mean burned area in years 2003-2013 & historical\_fires & v & \cite{ECMWF} \\ \hline
\multirow{6}{*}{fire weather indices} & fire weather index & fwi & v & \multirow{6}{*}{\cite{era5}} \\ 
 & buildup index & bui & v &  \\ 
 & drought code & drought & v &  \\ 
 & duff moisture code & duff & v &  \\ 
 & fine fuel moisture code & ffmc & v &  \\ 
 & initial spread index & isi & v &  \\ \hline
\multirow{3}{*}{spatiotemporal data} & latitude & lat & v & - \\ 
 & longitude & lon & v & - \\ 
 & month & month & v & - \\ \hline
\end{tabular}%
}
\end{table}

\subsection*{Dynamics analysis}

\begin{figure}[htp]
    \centering
    \includegraphics[width=0.99\textwidth]{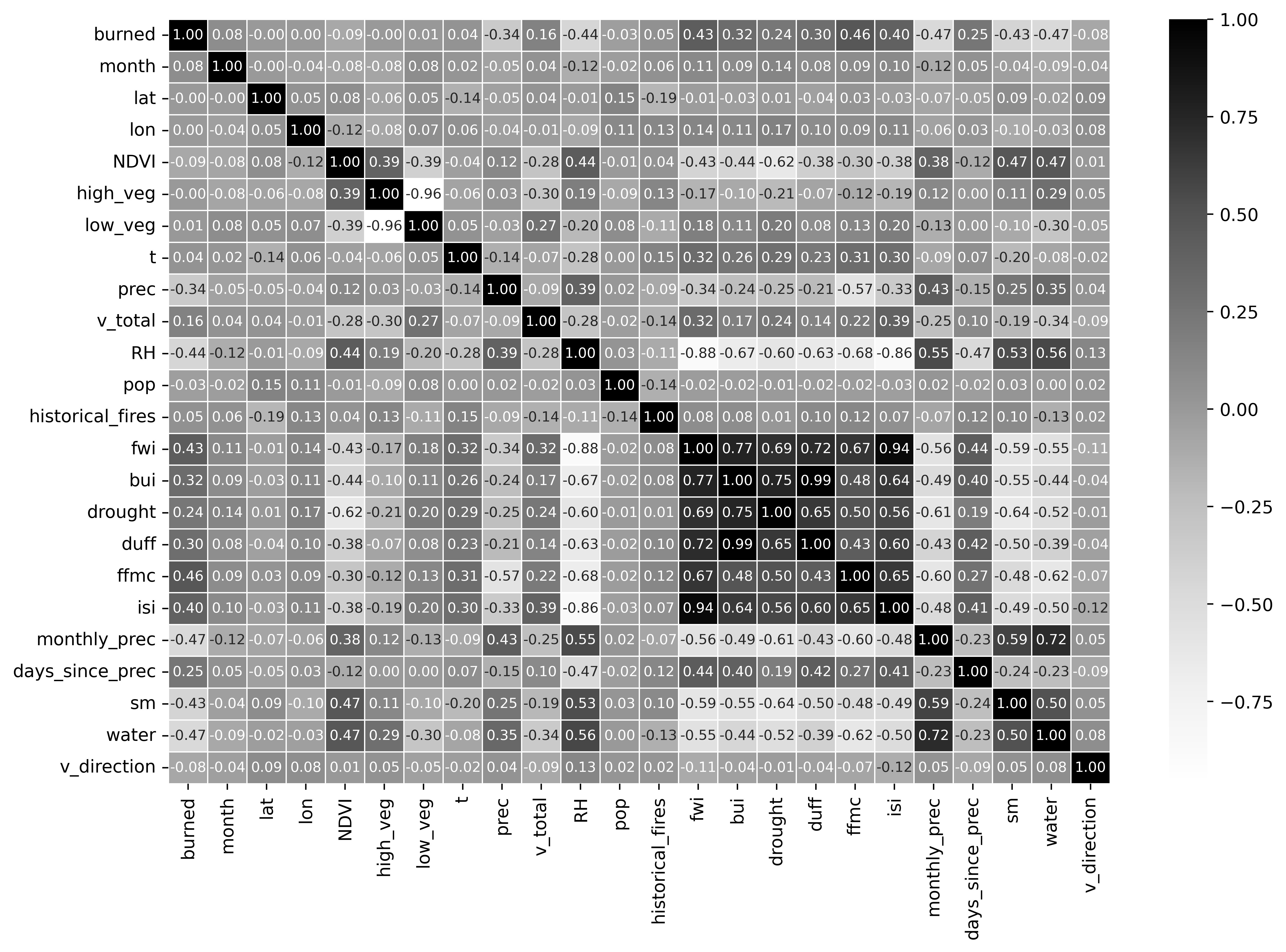}
    \caption{A Pearson correlation matrix between the source features and themselves and between the source features and the target feature.}
    \label{fig:pearsons}
\end{figure}

In order to appropriately decide on the modeling approach to the problem, we started by analyzing the dynamics that emerge from the data. We present the Pearson correlation matrix between the source features and themselves and between the source features and the target feature in Figure \ref{fig:pearsons}. Observing the data, it becomes apparent that the values in the first row are predominantly close to zero, with the highest absolute value reaching 0.48, indicating that the source features are not linearly related to the target features in a pair-wise manner. The features that are most strongly correlated to ignition probability are fire indices (fwi, ffmc) and Relative Humidity (RH). Moreover, as revealed in the figure, the source features are weakly linearly correlated to each other for most of the cases. Some exceptions are low and high vegetation cover, or fire weather indices which are correlated between themselves and some meteorological factors. These correlations are already known from previous studies \cite{compare_4}. This outcome indicates that the source space is mostly non-linear and an ML model can be useful for the proposed classification task. 

\begin{figure}[htp]
    \centering
    \includegraphics[width=0.99\textwidth]{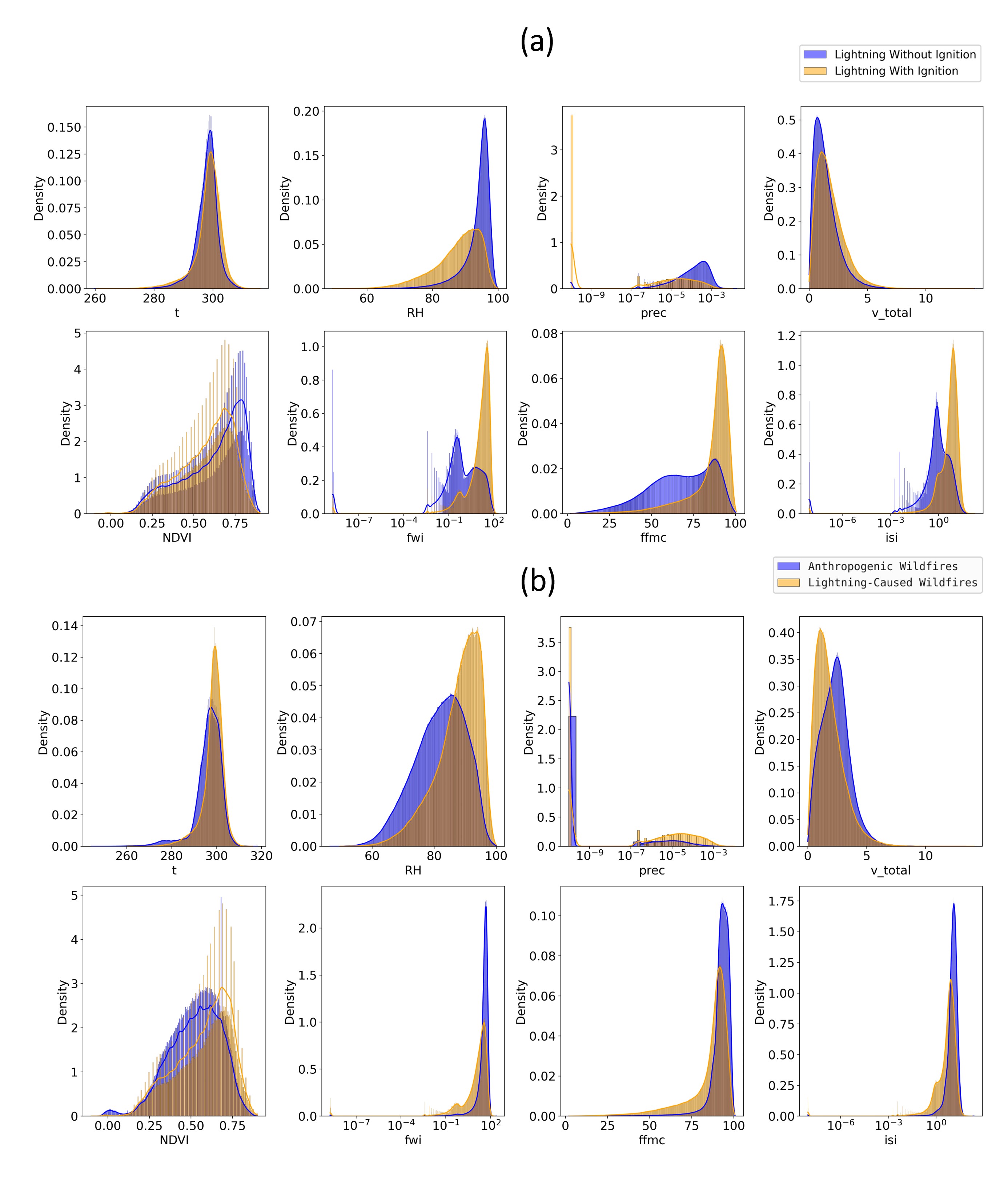}
    \caption{(a) Descriptive statistics of lightning with or without ignition (b) Descriptive statistics of anthropogenic versus lightning-ignited wildfires}
    \label{fig:combined_hists}
\end{figure}

In addition, Figure \ref{fig:combined_hists}a presents descriptive statistics of the obtained dataset and compares the different variables during thunderstorms where ignitions either did or did not occur. Eight selected variables are presented, and the remaining variables are presented in the Appendix (Figure S1).
Notably, the distributions of Relative Humidity (RH) and the Fine Fuel Moisture Code (FFMC) indexes, indicative of vegetation dryness, exhibit discernible differences between instances of ignitions and non-ignitions. This observation implies their potential value as critical variables in the predictive model.
Historical (monthly) precipitation, which is correlated with the ffmc index, also presents a different distribution between the two cases. Other variables such as temperature and wind velocity appear to have similar distributions for both cases.

\subsection*{Model's performance}
Table \ref{tab:ml_compare} shows the performance of four ML models - Logistic Regression, Random Forest, AutoGluon, and XGBoost for five different configurations of source features, divided into four sets - vegetation, meteorological, and anthropogenic factors, fire history, FWIs, and spatio-temporal data. The results for each feature configuration and model are shown as the accuracy of the obtained model on the test set. For all five feature configurations, a clear order of performance emerges where the XGboost model provides the highest accuracy followed by the AutoGluon, Random Forest, and Logistic Regression models. In particular, with all the features in the dataset, the XGboost obtains a 91.6\% accuracy, compared to the benchmark of a logistic regression with an accuracy of 82.1\%, a 9.5\% performance increase. Interestingly, comprising the fourth and fifth rows shows that the introduction of spatial data allows all models to highly improve their performance with XGboost increasing from 89.3\% to 91.6\%, a 3.7\% increase. Also, we find that including regional characteristics such as the historical burned area contributes to the performance of the model. The ROC curves and AUC scores are presented in Figure S4 in the Supplementary Information file. 

Figure \ref{fig:combined_maps}a and Figure \ref{fig:combined_maps}b present the mean regional accuracy of the XGBoost and logistic regression models, respectively. While the logistic regression performs relatively well in Africa, South America, and South Asia, it performs poorly in Australia, North Asia, and North America. In contrast, the XGBoost model exhibits excellent performance in all the continents, an indicator of good generalization capability. The accuracy of the model in the validation year, 2021, was almost identical, with 90.5\%  for the best-performing XGBoost model. These results indicate that the proposed model was able to generalize from the trained data and capture relevant patterns in the dynamics.

\begin{table}[h!]
\centering
\caption{Performance of the various models, differing by their machine learning model as well as the provided features.}
\label{tab:ml_compare}
\resizebox{\textwidth}{!}{%
\begin{tabular}{ccccccccccc}
\toprule
\multirow{2}{*}{Model \#} & \multicolumn{4}{c}{Features} & \multirow{2}{*}{Model Type} & \multirow{2}{*}{ROC AUC} & \multirow{2}{*}{Accuracy} & \multirow{2}{*}{F1 Score} & \multirow{2}{*}{Precision} & \multirow{2}{*}{Recall} \\ \cmidrule{2-5}
 & \multicolumn{1}{c}{\begin{tabular}[c]{@{}c@{}}Vegetation,\\ Meteorologic,\\ and Anthropogenic\\ Factors\end{tabular}} & \multicolumn{1}{c}{Fire History} & \multicolumn{1}{c}{\begin{tabular}[c]{@{}c@{}}Fire Weather\\ Indices\end{tabular}} & \begin{tabular}[c]{@{}c@{}}Spatio-\\ temporal\\ Data\end{tabular} &  &  &  &  &  &  \\ \midrule
\multirow{4}{*}{Model 1} 
 & \multicolumn{1}{c}{v} & \multicolumn{1}{c}{} & \multicolumn{1}{c}{} &  & Logistic Regression & 0.798 & 0.798 & 0.806 & 0.801 & 0.812 \\ \cmidrule{2-11} 
 & \multicolumn{1}{c}{v} & \multicolumn{1}{c}{} & \multicolumn{1}{c}{} &  & XGBoost & 0.877 & 0.877 & 0.881 & 0.880 & 0.882 \\ \cmidrule{2-11} 
 & \multicolumn{1}{c}{v} & \multicolumn{1}{c}{} & \multicolumn{1}{c}{} &  & Random Forest & 0.829 & 0.828 & 0.828 & 0.856 & 0.802 \\ \cmidrule{2-11} 
 & \multicolumn{1}{c}{v} & \multicolumn{1}{c}{} & \multicolumn{1}{c}{} &  & AutoGluon & 0.875 & 0.876 & 0.880 & 0.878 & 0.882 \\ \midrule
\multirow{4}{*}{Model 2} 
 & \multicolumn{1}{c}{v} & \multicolumn{1}{c}{v} & \multicolumn{1}{c}{} &  & Logistic Regression & 0.798 & 0.798 & 0.806 & 0.801 & 0.811 \\ \cmidrule{2-11} 
 & \multicolumn{1}{c}{v} & \multicolumn{1}{c}{v} & \multicolumn{1}{c}{} &  & XGBoost & 0.887 & 0.887 & 0.891 & 0.890 & 0.892 \\ \cmidrule{2-11} 
 & \multicolumn{1}{c}{v} & \multicolumn{1}{c}{v} & \multicolumn{1}{c}{} &  & Random Forest & 0.832 & 0.831 & 0.831 & 0.860 & 0.804 \\ \cmidrule{2-11} 
 & \multicolumn{1}{c}{v} & \multicolumn{1}{c}{v} & \multicolumn{1}{c}{} &  & AutoGluon & 0.885 & 0.886 & 0.890 & 0.888 & 0.892 \\ \midrule
\multirow{4}{*}{Model 3} 
 & \multicolumn{1}{c}{v} & \multicolumn{1}{c}{} & \multicolumn{1}{c}{v} &  & Logistic Regression & 0.805 & 0.805 & 0.812 & 0.811 & 0.812 \\ \cmidrule{2-11} 
 & \multicolumn{1}{c}{v} & \multicolumn{1}{c}{} & \multicolumn{1}{c}{v} &  & XGBoost & 0.885 & 0.885 & 0.889 & 0.888 & 0.890 \\ \cmidrule{2-11} 
 & \multicolumn{1}{c}{v} & \multicolumn{1}{c}{} & \multicolumn{1}{c}{v} &  & Random Forest & 0.831 & 0.830 & 0.831 & 0.856 & 0.808 \\ \cmidrule{2-11} 
 & \multicolumn{1}{c}{v} & \multicolumn{1}{c}{} & \multicolumn{1}{c}{v} &  & AutoGluon & 0.883 & 0.883 & 0.887 & 0.886 & 0.888 \\ \midrule
\multirow{4}{*}{Model 4} 
 & \multicolumn{1}{c}{v} & \multicolumn{1}{c}{v} & \multicolumn{1}{c}{v} &  & Logistic Regression & 0.805 & 0.805 & 0.811 & 0.812 & 0.811 \\ \cmidrule{2-11} 
 & \multicolumn{1}{c}{v} & \multicolumn{1}{c}{v} & \multicolumn{1}{c}{v} &  & XGBoost & 0.893 & 0.893 & 0.896 & 0.896 & 0.896 \\ \cmidrule{2-11} 
 & \multicolumn{1}{c}{v} & \multicolumn{1}{c}{v} & \multicolumn{1}{c}{v} &  & Random Forest & 0.835 & 0.834 & 0.835 & 0.860 & 0.811 \\ \cmidrule{2-11} 
 & \multicolumn{1}{c}{v} & \multicolumn{1}{c}{v} & \multicolumn{1}{c}{v} &  & AutoGluon & 0.891 & 0.891 & 0.895 & 0.896 & 0.894 \\ \midrule
\multirow{4}{*}{Model 5} 
 & \multicolumn{1}{c}{v} & \multicolumn{1}{c}{v} & \multicolumn{1}{c}{v} & v & Logistic Regression & 0.821 & 0.821 & 0.827 & 0.827 & 0.827 \\ \cmidrule{2-11} 
 & \multicolumn{1}{c}{v} & \multicolumn{1}{c}{v} & \multicolumn{1}{c}{v} & v & XGBoost & \textbf{0.916} & \textbf{0.916} & \textbf{0.919} & \textbf{0.916} & \textbf{0.922} \\ \cmidrule{2-11} 
 & \multicolumn{1}{c}{v} & \multicolumn{1}{c}{v} & \multicolumn{1}{c}{v} & v & Random Forest & 0.846 & 0.846 & 0.848 & 0.867 & 0.829 \\ \cmidrule{2-11} 
 & \multicolumn{1}{c}{v} & \multicolumn{1}{c}{v} & \multicolumn{1}{c}{v} & v & AutoGluon & 0.913 & 0.913 & 0.916 & 0.914 & 0.919 \\ \bottomrule
\end{tabular}%
}
\end{table}

Figure~\ref{fig:combined_maps}c-g presents the estimated risk for lightning-ignited wildfires. We present both annual mean values (Figure~\ref{fig:combined_maps}c) and seasonal averages (Figures~\ref{fig:combined_maps}d-g).  There is a difference in the distribution of lightning-ignited wildfires in the tropics compared with high latitudes. In the tropics, lightning fires generally occur in the dry season, or the transition between seasons, when the vegetation is dry, but there can still be isolated thunderstorms that can ignite the vegetation. In high latitudes, such as the boreal forests of Canada, Alaska, and Siberia, lightning fires always occur in the summer months that are associated with thunderstorms. While thunderstorms are also associated with rainfall, the rain falls directly below the clouds, while the lightning can strike tens of kilometers away from the rainfall.  Furthermore, after a dry period without rainfall, forests can ignite from lightning even in the presence of rainfall.  While it is generally thought that tropic lightning fires are rare, due to the high moisture in the tropics, this study shows a significant risk for lightning fires in the tropics throughout the year.

\begin{figure}[htp]
    \centering
    \includegraphics[width=14cm]{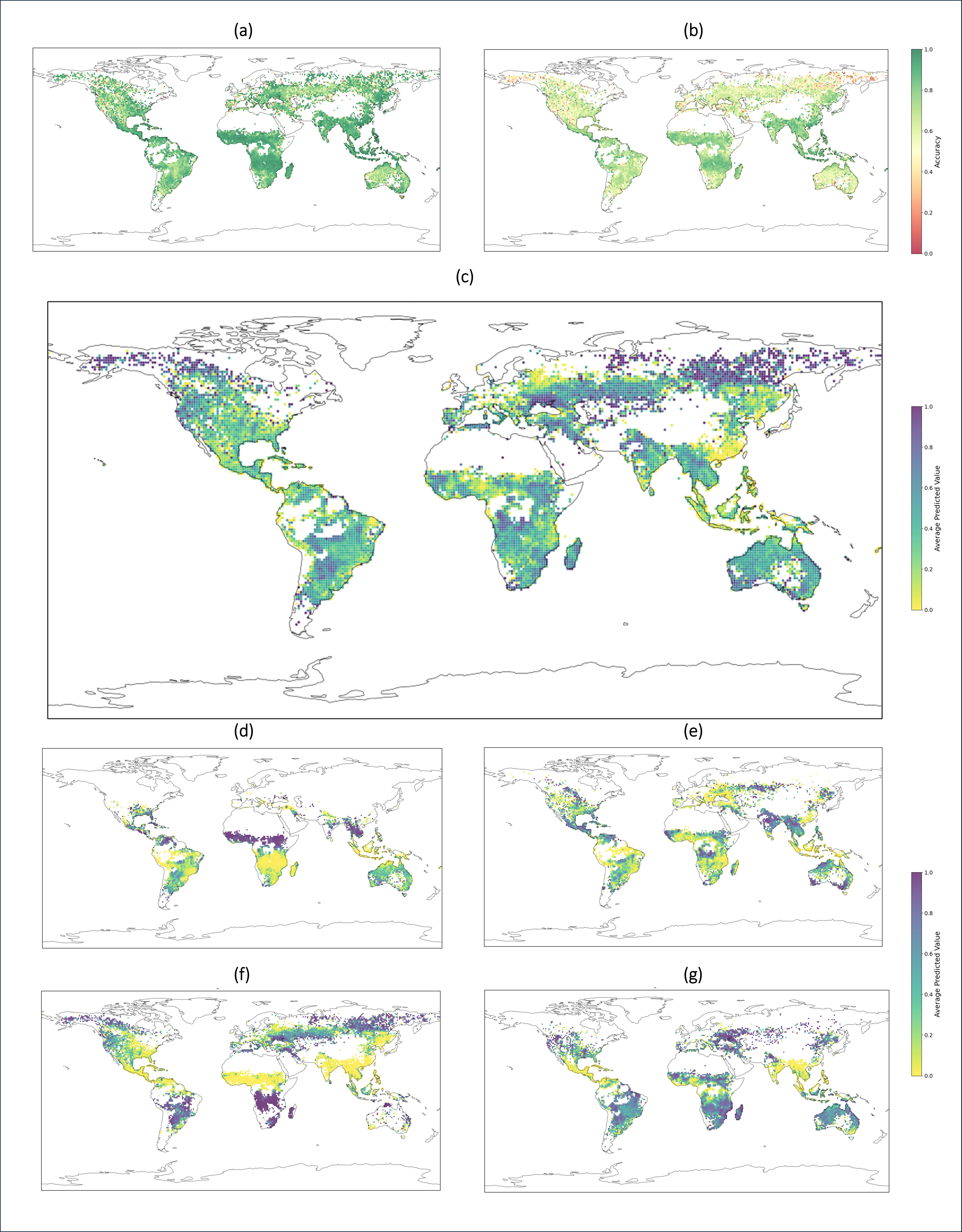}
    \caption{Model accuracy and predicted seasonal ignition risk based on the full XGBoost model. (a) Average accuracy of the XGBoost model (b) Average accuracy of the logistic regression (c) Predicted annual ignition risk based on the full XGBoost model (d-g) Predicted seasonal ignition risk based on the full XGBoost model: (d) December-February (e) March-May (f) June-August (g) September-November. Similar figures for additional models are presented in the Supplementary Information file (Figures S5-S7). The figure was generated using the Cartopy library \cite{cartopy}.}
    \label{fig:combined_maps}
\end{figure}

\subsection*{Anthropogenic versus lightning-ignited wildfires}

We present descriptive statistics of anthropogenic versus lightning-ignited wildfires in Figure~\ref{fig:combined_hists}b. Eight selected variables are presented, and the remaining variables are presented in the Appendix (Figure S3). The histograms in the figure describe the different variables on the day of ignition for both wildfire types. The histograms highlight some important differences between the two wildfire types: First, the meteorological conditions are remarkably different, with lightning ignitions occurring in high RH values and nonzero precipitation. Obviously, this does not mean that high RH or precipitation caused lightning-ignitions, but rather a description of the typical meteorological factors during thunderstorms. The typical wind velocity is small in lightning-ignitions; it is possible that strong winds at ground level are an indicator of strong winds at higher altitudes, which hinder the accumulation of thunderclouds. Second, and with a direct link to the meteorological factors, the fire weather indices in lightning-ignited ignitions are often very low. This is not surprising, as these indices are a function of the meteorological conditions, and high RH and precipitation normally indicate a low risk of ignitions. Finally, the vegetation indices are also different between the two ignition causes; lightning-ignited ignitions occur in regions with higher NDVI values.

These distinctively different characteristics suggest that the prediction of anthropogenic and lightning-ignited wildfires requires separate models, as these are two different phenomena. To examine this issue, we trained an XGBoost model to predict anthropogenic wildfires and tested its performance in the lightning-ignited wildfires dataset, and vice versa. The results, as summarized in Table \ref{tab:cross-fire-type-a} and Table \ref{tab:cross-fire-type-b}, confirm the inadequate performance of both models in predicting wildfires originating from an ignition source different from the one on which they were trained. In similarity to the XGBoost model, fire weather indices could also be adapted to reflect the risk of lightning-ignited ignitions.

\begin{table}[!ht]
  \label{tab:cross-fire-type}
  \caption{Prediction of anthropogenic versus lightning-ignited wildfires}
  \begin{subtable}{0.5\linewidth}
  \centering
  \caption{Basic Model (no geo-location)}
  \label{tab:cross-fire-type-a}
  \begin{tabular}{lcc}
    \toprule
    \textbf{Training \textbackslash Testing} & \textbf{Anthropogenic} & \textbf{lightning-ignited} \\
    \midrule
    \textbf{Anthropogenic} & 82.2\% & 70.6\% \\
    \textbf{lightning-ignited} & 63.2\% & 89.3\% \\
    \bottomrule
  \end{tabular}
  \end{subtable}%
  \begin{subtable}{0.5\linewidth}
  \centering
  \caption{Full Model}
  \label{tab:cross-fire-type-b}
  \begin{tabular}{lcc}
    \toprule
    \textbf{Training \textbackslash Testing} & \textbf{Anthropogenic} & \textbf{lightning-ignited} \\
    \midrule
    \textbf{Anthropogenic} & 86.5\% & 74.9\% \\
    \textbf{lightning-ignited} & 72.2\% & 91.6\% \\
    \bottomrule
  \end{tabular}
  \end{subtable}% 
\end{table}

\subsection*{Climate change influence and projections}
\begin{figure}[!ht]
    \centering
    \includegraphics[width=14cm]{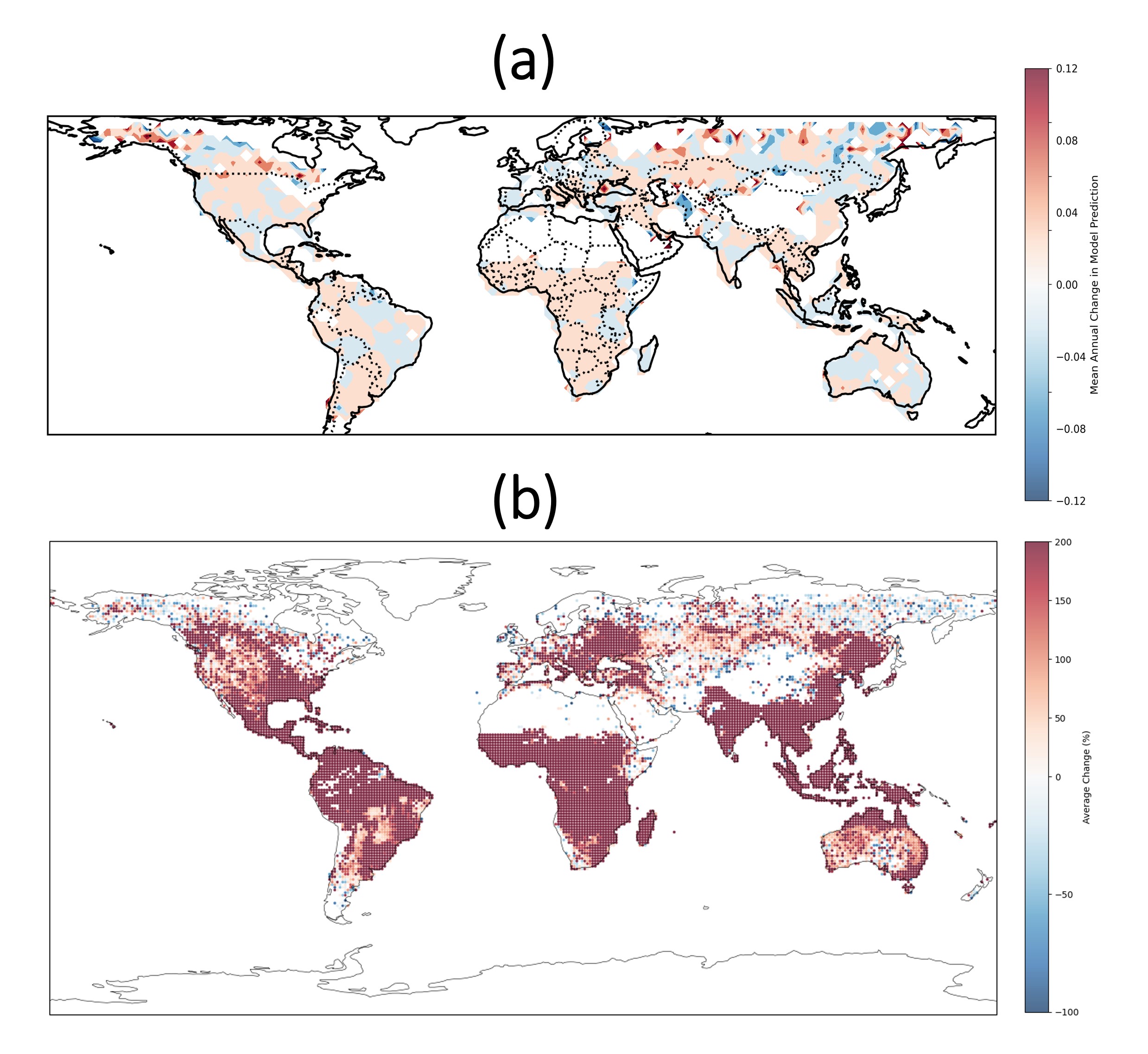}
    \caption{(a) Mean annual difference in the model's prediction of ignition risk between 2014-2021. (b) Increase in wildfire risk between present day and 2100. The figure was generated using the Cartopy library \cite{cartopy}.}
    \label{fig:climate-change}
\end{figure}

Figure \ref{fig:climate-change}a presents the mean annual difference for the model's prediction throughout the eight years of the data's timespan (2014-2021). When averaging over the entire globe, we find a mean of 1\% and a median of 0.7\% annual increase in lightning-caused wildfire risk. However, this increase is not uniformly distributed in all regions. The increase in Africa, South America, and the USA is small but relatively consistent throughout these regions. In contrast, the trends in Australia, Canada, and North Asia is mixed and region-dependent. The possible decrease in ignition risk at high latitudes is consistent with previous studies, which have recognized increased precipitation as a mitigating factor of lightning-caused ignitions \cite{compare_2}.
Figure \ref{fig:climate-change}b presents the increase in lightning ignition risk between present day and the projected risk for 2100. The model predicts an increase of 50\% in lightning ignition risk. Although these two analyses are distinctively different, the regional distributions of their outcomes are remarkably similar. The lightning ignitions risk is consistently increasing in Africa, South America, and the USA, in contrast to mixed patterns in Canada and North Asia. Unlike the analysis in Figure \ref{fig:climate-change}a, climate projections point to a clear increase in Australia.

\section*{Discussion}
\label{sec:methods}
We created a dataset that classifies anthropogenic and lightning-ignited wildfires globally based on high-resolution spatio-temporal observations of wildfires and thunderhours. We developed ML models to predict the risk of ignition when lightning occurs and obtained a high predictive performance of 91.6\% using the XGboost model. The model was trained on a global dataset and based on common and easily available variables, making it applicable everywhere. The high performance of the model was also validated on data from 2021 as a hold-out year that was not used for training, suggesting this model generalizes the patterns well and could be applicable in future applications.

We analyze the most important features in the best performing models in the Supplementary Information file. Figure S8 presents an analysis of the most important features in the XGBoost model in the form of Shapley values. Precipitation on the day of ignition is the most important factor and is negatively correlated with ignition risk, although ignitions do occur with limited precipitation during thunderstorms. The next most important feature is the fire weather index, which is mostly positively correlated with ignition risk despite a non-negligible number of low fwi values which were still associated with high ignition risk. RH, ffmc, and monthly precipitation all had the expected impact on the model's outcome. Temperature had a relatively small effect on the model. Its small, negative correlation with ignition risk is likely a spurious result, possibly influenced by the regions where thunderstorms commonly occur. Similarly, Figure S9 presents dependence plots for the ten most important features of the XGBoost model based on the SHAP analysis. The impact of some features on the model aligns with expectations. For instance, ffmc, RH, soil moisture, and vegetation water content values exhibit a nearly linear and strong effect on wildfire risk. The influence of other features, however, is not linear. For example, wildfire risk increases with NDVI values up to approximately 0.8 but then decreases, which might be due to very high NDVI values indicating wet and healthy vegetation that is less flammable. Fire history also shows a mixed effect on the model. Very low fire history values suggest a lower risk of wildfire ignition, while very high values also appear to have a negative effect, possibly because previous fires have already exhausted the available flammable vegetation. The fire weather index has a positive effect on the model, demonstrating that traditional indices have relatively good predictive capability. Figure S10 shows the ranked importance of features according to the XGBoost model. As expected, the one-hot encoded month variables significantly impact the model. Other key features include daily and monthly precipitation, soil moisture, NDVI, and others. This analysis aligns with the SHAP feature importance analysis, reinforcing its conclusions.

We demonstrated that anthropogenic and lightning-ignited wildfires have distinctively different characteristics and commonly occur at different RH, precipitation, and other conditions. In fact, as lightning-ignited wildfires can only occur following a lightning stroke, the meteorological conditions in which they begin are systematically biased compared to anthropogenic wildfires which can ignite without thunderstorms. This dramatically reduces the predictive performance of models trained on anthropogenic wildfires when applied to lightning-ignited ignitions, and vice versa. We demonstrated this finding by training an ML model for anthropogenic wildfires and testing it for lightning-ignited wildfires, which reduced its accuracy to as low as 75\%. This highlights the need to address anthropogenic and lightning-ignited wildfires as different phenomena and develop distinct prediction models and fire weather indices for the two ignition sources.

Based on the developed model, we analyzed the influence of climate change on wildfire ignition risk throughout the years. We did so both by the annual predicted ignition risk in our dataset (2014-2021), and by applying the model to climate change projections for 2100. Both analyses provide a similar conclusion - lightning-ignited wildfires are already rising in Africa, South America, and the USA, and will likely keep rising substantially following climate change. In contrast, the trend in Australia, Canada, and North Asia is not as clear, though regional variations are expected in these areas as well.

This study provides a novel approach to prediction of lightning-caused wildfires, but it does have some noteworthy limitations. First, although the classification of anthropogenic versus lightning-caused wildfires is performed at a high resolution, the uncertainty in holdover fire duration prevents a certain classification. Namely, our classification assumes that a wildfire that ignites in the week after the thunderstorm is caused by it. Furthermore, we chose to analyze the effect of climate change in two ways: either based on our data (2014-2021), or based on climate change projections. The analysis based on the actual data makes its conclusion solid and robust, but it comes at the price of basing our findings regarding climate change on eight years of data, which is not necessarily sufficient for this matter. In contrast, using the climate projections provides insights on a longer time frame, but is not as accurate as the empirical analysis. Future studies could elaborate this analysis both for the past (by using alternative lightning datasets) or by elaborating the climate projections analysis. Future studies could also address the challenge of improving the prediction and detection of holdover fires, which could potentially serve as an early warning for large wildfires. Managing wildfires during the smoldering phase is considerably easier, making timely detection a crucial opportunity for effective wildfire mitigation. Finally, while the current study focused on ignition risk during a thunderstorm, to fully comprehend future lightning-ignited wildfires one must also consider the frequency of thunderstorms under climate change. 

The accurate ML model presented in this study holds the potential to augment existing firefighting strategies. With the escalating risk of wildfires triggered by lightning, an enhanced understanding and predictive model serve as crucial initial steps for effective fire mitigation. Accurate prediction models are equally vital for promptly alerting nearby populations to imminent dangers. To fulfill this purpose, we advocate for the establishment of dedicated fire weather indices for lightning ignitions, leveraging insights from the model developed in this study.

\section*{Methods and Materials}
\begin{figure}[!ht]
    \centering
    \includegraphics[width=0.99\textwidth]{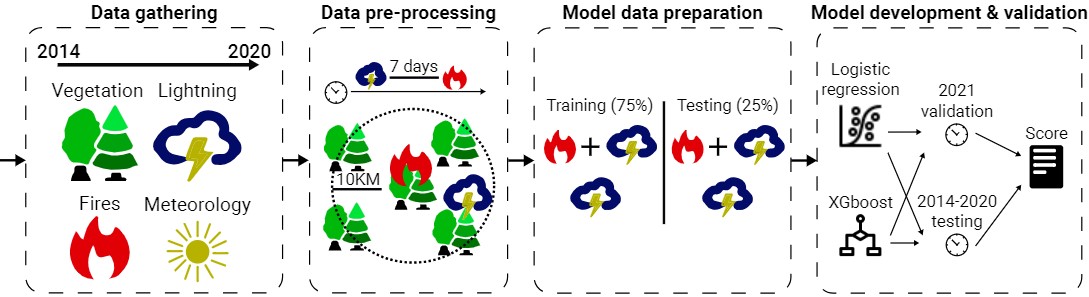}
    \caption{A schematic view of the methodology used in this study.}
    \label{fig:scheme}
\end{figure}

This section presents the three main components of constructing this study: the dataset, the ML model of lightning-ignited fire prediction, and the performed experiments. Initially, we outline the open-access datasets taken into consideration as well as their preparation and integration into a single dataset for the ML model. Subsequently, we formally describe the proposed ML model training and validation processes. Finally, based on both the datasets and the proposed ML model, we describe several experiments designed to approximate the effect of climate change on lightning-ignited fire on a global scale. Fig. \ref{fig:scheme} presents a schematic view of the methodology used in this study.

\subsection*{Datasets}

\subsubsection*{Wildfire data}
Wildfire data are taken from \cite{arteswildfires}. This dataset classifies wildfires to the individual wildfire level with global coverage and includes shape files representing the daily burned area for each wildfire in each day, at 250-meter resolution. For each wildfire, we find the center of the polygon describing the burned area on its first day and define it as the ignition point. We do so for all wildfires in the years 2014-2021.

\subsubsection*{Lightning data}
\begin{comment}
%this is original text. Replaced below with suggestion by Collin. 
To classify lightning-ignited wildfires, we obtained global thunderhours coverage for the study's timespan from \cite{thunderhours} provided in a 0.05-degree resolution. The dataset is derived from the Earth Networks Global Lightning Detection Network; the use of ground-based systems that are globally distributed improves the spatial coverage of lightning detection including higher latitudes, though at some cost of detection efficiency. For this reason, we use thunder hours instead of individual lightning observations, which are substantially preferable in terms of detection efficiency. Thus, we use data which is both global and of relatively high detection efficiency \cite{poreba}. 
\end{comment}
To classify lightning-ignited wildfires, we have used the thunder hours data recently produced from the Earth Networks Total Lightning Detection Network (ENTLDN) \cite{thunderhours}.  This ground-based lightning detection network observes the electromagnetic pulses radiated from lightning discharges, and using hundreds of stations around the globe can geolocate the lightning discharges in time and space.  However, all lightning networks have problems with detection efficiencies since they are biased to the most intense flashes, while missing many of the weaker flashes. To overcome this deficiency, ENTLDN have gridded the lightning data into a global 5 km horizontal grid. If at least 2 flashes are detected within a 5 km gridbox during a particular hour, that gridbox is given the value of 1, otherwise 0. In this way instead of counting lightning flashes, ENTLDN counts thunderstorms per hour, or how many hours an observer would have heard thunder at that location. These thunderhous can be counted per day, month, or year for a specific location. Hence, our study uses thunderhour as the parameter representing the risk of lightning in a particular location and at a particular time.

\subsubsection*{Meteorological data and vegetation}
We use ERA5 reanalysis \cite{era5} for historical meteorological data. We include daily means of temperature, relative humidity, precipitation, and wind data. As studies show that time-lagged meteorological factors have a substantial impact on wildfire risk \cite{historicalmet}, we also include two additional features: mean precipitation in the month prior to each observation, and the total days since the last precipitation.
We also include three vegetation-related features: we obtain low vegetation cover and high vegetation cover from ERA5 reanalysis \cite{era5}; we use the normalized difference vegetation index (NDVI), an estimate of the density of live green vegetation, obtained from the NASA Earth Observations website \cite{NDVI}. Finally, we include vegetation water content (skin reservoir) and soil moisture from the ERA5 reanalysis \cite{era5}.

\subsubsection*{Historical burned areas}
We include a feature describing the mean historical burned area in the years 2003-2013. For this feature, we do not include data from 2014 onwards to avoid data leakage. The historical burned areas are obtained from the ECMWF website \cite{ECMWF} in 0.25-degree resolution.

\subsubsection*{Fire weather indices}
We use the Canadian Forest Fire Weather Index System \cite{canadianfwi} which consists of six indices that account for the effects of fuel moisture and weather conditions on fire behavior: fine fuel moisture code, duff moisture code, drought code, initial spread index, buildup index, and the fire weather index which serves as a general danger index built on the previously mentioned ones. We obtain these variables in daily 0.25° resolution from the Copernicus Climate Change Service \cite{copernicus}. Although these indices are based on meteorological data, research shows that integrating significant features based on domain knowledge can improve machine learning performance \cite{sr}.

\subsubsection*{Dataset integration}
For each wildfire in the data, we extracted its ignition point and date. We classified a wildfire as lightning-ignited if thunderhours occurred in a range of 10 kilometers on the day of ignition, or the week before it (to account for holdover wildfires). We eliminated wildfires which lasted for a single day to avoid prediction of agricultural burning. We then balanced the data by randomly sampling a similar number of observations in which a thunderhour occurred, but no wildfires ignited within a week in a range of 10 kilometers. We left the year 2021 as a hold-out year, meaning it was not used for training at all, and only served for validation purposes.

At this stage, the dataset (\(D\)) is represented as a single table with \(n = 974459\) positive samples (lightning-ignited wildfires) and \(m = 18\) features. Formally, each sample contains \(\) source features (\(x\)) and the target feature (\(y\)) which indicates if a fire is lightning ignited or not. This structure dictates a binary classification task. To this end, we first computed the Pearson coefficient between each pair of features in the dataset (including the target feature). Afterwards, since these results reveal that there is no significant linear or statistical connection between each feature to the target feature as well as a non-linear connection between the source features, we decided to use ML models for this classification task \cite{classificaiton_ml_review}. 

To this end, we first pre-process the dataset to a form that is usable by ML models. Namely, we first removed all positive samples where at least a single datapoint is missing - resulting in a removal of \(1.6\%\) of the samples. Moreover, as the dataset is unbalanced (i.e., in the vast majority of cases lightnings do not result in wildfires), we decided to apply a down-sampling strategy. Namely, in addition to the lightning-ignited wildfires, we included an equal number of lightning observations that did not result in ignitions, sampled at random. The global distribution of lightning-caused and anthropogenic wildfires is presented in Figure S2.

\subsection*{Machine learning model development}

In order to find the ML model that produces the highest accuracy score on the test set, we tested multiple popular ML models, including Logistic Regression (LR) \cite{logistic_regression}, Random Forest (RF) \cite{rf}, AutoGluon \cite{autogluon}, and XGboost \cite{xgboost}. We chose these models as previous studies show these models obtain promising results in similar spatiotemporal geophysical tasks \cite{ml_geo_1,ml_geo_2}. To be exact, LR is the classification version of the popular linear regression \cite{linear_regreesion} model where one assumes the relationship between the source and target features is linear. The RF model is an ensemble ML model that generates multiple decision tree (DT) models by providing a different and randomly sampled subset of the samples and features of the dataset to each DT and makes a final prediction using the majority-vote aggregation method. The XGBoost classifier, or eXtreme Gradient Boosting, is a robust ML technique that enhances the performance of DTs through an iterative process \cite{xgboost}. In contrast to traditional DTs, XGBoost constructs a sequence of trees, each aiming to correct the errors made by the previously trained trees. This sequential training is guided by the principles of gradient boosting. Finally, AutoGluon is an Automatic Machine Learning (AutoML) library that uses ensemble models of Machine and Deep Learning \cite{autogluon}.

On top of that, as deep learning models obtain promising results in a wide range of fields \cite{teddy_drl_2023,dl_2,dl_3}, in general, and in geophysics, in particular \cite{dl_geo}, we take advantage of the AutoGluon library \cite{autogluon}, an Automatic Machine Learning (AutoML) library which uses ensemble models of Machine and Deep Learning. This tool uses a unique search and optimization process to find and test a large number of neural network models as well as tree-based ensemble models, looking to optimize the model's accuracy on the test set.  

\subsection*{Experimental setup}
We develop several models based on different features and different ML techniques (as presented in Table \ref{tab:ml_compare}). The features included in the model vary from the basic model, which only includes meteorological data, population density, and vegetation data, to the full model which additionally includes regional fire history, fire weather indices, and spatiotemporal data. We present the accuracy scores of the four different models: logistic regression, XGBoost, Random Forest, and AutoGluon.

In Figure \ref{fig:combined_maps}a,b we present the accuracy of the best performing model, XGBoost, and logistic regression as a benchmark. Both models are based on the full set of features. Figure \ref{fig:combined_maps}c presents the annual average of the XGBoost model's prediction in different regions, and \ref{fig:combined_maps}d-g provides the seasonal average of its prediction.

We evaluate the performance of the model, which is trained on lightning-caused wildfires, when tested on anthropogenic wildfires, and vice versa. To do so, we train an additional XGBoost model on all the wildfires in the dataset which were not classified as lightning-caused. In addition, we randomly chose a similar number of random observations in which there were no ignitions (and no lightning). We trained an additional XGBoost model for anthropogenic wildfires, and evaluated either model on either dataset. We repeated this process twice, using either the basic set of features or the full model.

To estimate the effect of climate change on lightning ignition risk, we perform two analyses. In the first analysis (presented in Figure \ref{fig:climate-change}a), we evaluate the mean annual change in the model's prediction. We randomly selected eight million thunder samples from the eight years of data (2014-2021). We aggregated samples that are located in similar regions, and calculated the mean model prediction in each region, in each year. We then calculated the mean annual change in each region. Next, we applied the model to climate change projections for 2100 based on the EC-Earth3-CC model under SSP245 scenario \cite{CMIP6} (presented in Figure \ref{fig:climate-change}b). Inspired by the methodology of previous studies \cite{compare_2}, we added the mean projected differences between 2100 and present-day for three main meteorological factors (RH, temperature, and precipitation), while holding the rest of the variables constant. We present the difference in the model's prediction between the two.

\bibliography{main}

\section*{Acknowledgements}
We acknowledge the World Climate Research Programme, which, through its Working Group on Coupled Modelling, coordinated and promoted CMIP6. We thank the climate modeling groups for producing and making available their model output, the Earth System Grid Federation (ESGF) for archiving the data and providing access, and the multiple funding agencies that support CMIP6 and ESGF. We also thank the Earth Networks Lightning Network for making their novel thunderhour data available.

\section*{Author contributions statement}
Conceptualization: A. S., C. P.; software: A. S., T. L.; Validation: T. L., O. G., C. P.; Formal Analysis: A. S., O. G., T. L.; data curation: A. S.; Writing - Original Draft: A. S., T. L.; Visualization: A. S., T. L.; Supervision: T. L., O. G., E. H., C. P.; Project administration: A. S.; All authors designed the methodology, investigated, and reviewed the manuscript. 

\section*{Ethics declarations}
The authors declare no competing interests.

\section*{Data availability}
The datasets generated during and/or analyzed during the current study are available from the corresponding author upon reasonable request.

\end{document}